\documentclass[11pt]{article}
\usepackage{coling2016}
\usepackage{times}
\usepackage{url}
\usepackage{latexsym}
\usepackage{amsmath}
\usepackage{amssymb}
\usepackage{graphicx}
\usepackage{color}

\newcommand{\ov}{\overrightarrow}
\newcommand{\ovl}{\overline}

\newcommand{\F}{{\cal F}}

\title{Distributional Inclusion Hypothesis for Tensor-based Composition}

 \author{Dimitri Kartsaklis \and Mehrnoosh Sadrzadeh \\
         Queen Mary University of London \\ 
         School of Electronic Engineering and Computer Science \\
         Mile End Road, London, UK \\
         {\tt\small \{d.kartsaklis;mehrnoosh.sadrzadeh\}@qmul.ac.uk}
         }

\date{}

\begin{document}

\maketitle

\blfootnote{
    %
    %
    %
    %
    \hspace{-0.70cm}  
    This work is licensed under a Creative Commons 
    Attribution 4.0 International Licence.
    Licence details:
    \url{http://creativecommons.org/licenses/by/4.0/}
    %
    %
}

\begin{abstract}
According to the distributional inclusion hypothesis,   entailment between  words can be measured via the feature inclusions of their  distributional vectors. In recent  work, we  showed how this hypothesis  can be extended  from words to   phrases and sentences in the setting of  compositional distributional semantics.    This paper focuses on inclusion properties of   tensors;  its main contribution is a theoretical and experimental analysis of how feature inclusion works in different concrete models of verb tensors. We present results for   relational, Frobenius,  projective, and holistic  methods and compare them  to the simple  vector addition, multiplication, min, and max models.    The degrees of entailment thus obtained  are evaluated   via a variety of existing word-based measures,  such as Weed's and Clarke's, KL-divergence, APinc,  balAPinc, and  two of our previously  proposed metrics at the phrase/sentence level. We perform experiments on three entailment datasets, investigating which  version of tensor-based composition achieves the highest performance when combined with the sentence-level measures.
\end{abstract}

\section{Introduction}
\label{sec:intro}

Distributional hypothesis asserts that   words that often occur in the same contexts have similar meanings \cite{firth1957}. Naturally these models are used extensively to measure  the semantic similarity of words \cite{turney2010}. Similarity is an a-directional  relationship and  computational linguists are  also interested in measuring degrees of directional relationships between words.   Distributional inclusion hypothesis is exactly about such relationships, and particularly, about entailment \cite{Dagan1999,geffet2005,herbelot2013}. It states that a word $u$ entails a word $v$ if in any context that   word  $u$ is used so can be word $v$. For example, in a corpus of   sentences `a boy runs',  `a person runs',   `a person sleeps', according  to this hypothesis,    {\em boy} $\vdash$ {\em person}, since wherever `boy' is used, so is `person'. Formally, we have that  $u$ entails $v$ if features of $u$ are included  in features of $v$, where features are non-zero contexts. In this example,  {\em boy} $\vdash$ {\em person},  since  $\{$run$\} \subset \{$run, sleep$\}$. 


For the same reasons that the   distributional hypothesis is not directly applicable  to phrases and sentences, the distributional inclusion hypothesis   does not scale up from words to larger language constituents. In a nutshell, this is because the majority of phrases and sentences  of language do not frequently occur in corpora of text, thus   reliable statistics  cannot be collected for them. In distributional models,   this problem is addressed with the provision of composition operators, the purpose of which is to combine the statistics of words and obtain statistics for   phrases and  sentences. In recent work, we have applied the same compositionality principles to  lift the entailment relation from word level  to phrase/sentence level \cite{lacl2016,isaim2016,Balkir2016,balkir2014}. The work in  \cite{balkir2014,Balkir2016} was focused on the use of  entropic measures on  density matrices and compositional operators thereof, but no experimental results were considered; similarly, \newcite{bankova2016} use a specific form of density matrices to represent words for entailment purposes, focusing only on theory. In \cite{isaim2016}, we showed how entropic and other measures can be used on vectors  as well as on density matrices and supported this claim with experimental results. In \cite{lacl2016}, we focused on making the distributional inclusion hypothesis compositional, worked out   how   feature inclusion lifts from words to compositional operators on them,  and based on experimental results showed that intersective composition operators result in more reliable entailments. This paper takes a more concrete perspective and focuses on the feature inclusion properties  of   tensors  constructed in different ways and composition operators applied to the tensors and vectors. 


One can broadly classify  the  compositional distributional models to three categories: ones that are based on simple element-wise operations between vectors, such as addition and multiplication \cite{lapata2010}; tensor-based models in which relational words such as verbs and adjectives are tensors and matrices  contracting and multiplying with noun (and noun-phrase) vectors \cite{CoeckeSadrClark2010,grefenstette2011,baroni2010}; and models in which the compositional operator is part of a neural network \cite{socher2012,nal2014} and is usually optimized against a specific objective. 

Tensor-based models stand in between the two extremes of element-wise vector mixing and neural net-based methods, offering a sufficiently powerful alternative that allows for theoretical reasoning at a level deeper than it is usually possible with black-box statistical approaches. Models of this form have been used in the past with success in a number of NLP tasks, such as head-verb disambiguation \cite{grefenstette2011}, term-definition classification \cite{KartSadrPul-COLING-2013}, and generic sentence similarity \cite{kartsadrqpl2014}. In this paper we extend this work to entailment, by investigating (theoretically and experimentally) the properties of feature inclusion  on the phrase and sentence vectors produced in four different  tensor-based compositional distributional models: the relational models of \cite{grefenstette2011}, the Frobenius models of \cite{KartSadrPul-COLING-2013,milajevs2014},   the projective models of \cite{lacl2016}, and the holistic linear regression model of \cite{baroni2010}.

Contrary to formal semantic models, and customary to distributional models,   our entailments are non-boolean and come equipped with degrees. We  review a number of measures that have been developed   for evaluating degrees of entailment at the lexical level, such as the \textit{APinc} measure and  its newer version, \textit{balAPinc} of \newcite{kotlerman2010}, which is considered as  state-of-the-art for word-level entailment. A newly proposed adaptation of these metrics, recently introduced by the authors in \cite{lacl2016}, is also detailed. This measure takes into account the specificities introduced by the use of a compositional operator and lifts the measures from words to  phrase/sentence level. 

We experiment with these models and evaluate them on  entailment relations between simple intransitive sentences, verb phrases, and transitive sentences on  the  datasets of \cite{lacl2016}. Our findings suggest that the  Frobenius models provide the highest performance, especially when combined with our sentence-level measures. On a more general note, the experimental results of this paper  support that of previous work \cite{lacl2016} and strongly indicate that compositional models employing some form of intersective feature selection, i.e. point-wise vector multiplication or tensor-based models with an element of element-wise mixing (such as the Frobenius constructions), are more appropriate for entailment tasks in distributional settings.


\section{Compositional distributional semantics}
\label{sec:compdist}

The purpose of a compositional distributional model is to produce a vector representing the meaning of a phrase or a sentence by combining the vectors of its words. In the simplest case, this is done by element-wise operations on the vectors of the words \cite{lapata2010}. Specifically, the vector representation $\ov{w}$  of a sequence of words ${w}_1, \dots, {w}_n$ is defined to be:

\[
\ov{w} := \overrightarrow{w_1}  +   \overrightarrow{w_2} + \cdots +  \overrightarrow{w_n} \hspace{2cm}
\ov{w} := \overrightarrow{w_1}  \odot   \overrightarrow{w_2} \odot \cdots \odot  \overrightarrow{w_n }
\]

In a more linguistically motivated approach, relational words such as verbs and adjectives are treated as linear or multi-linear maps. These are then  applied to the vectors of their arguments by following the rules of the grammar \cite{CoeckeSadrClark2010,grefenstette2011,baroni2010}. For example, an adjective is a map $N \to N$, for $N$ a basic noun space of the model.   Equivalently, this map can be represented  as a matrix  living in the space $N\otimes N$. In a similar way, a transitive verb is a map $N\times N \to S$, or equivalently, a ``cube'' or a tensor of order 3 in  the space  $N \otimes N \otimes S$, for $S$ a basic sentence space of the model. Composition takes place by tensor contraction, which is a generalization of matrix multiplication to higher order tensors. For the case of an adjective-noun compound, this simplifies to matrix multiplication between the adjective matrix and the vector of its noun, while for a transitive sentence it takes the following form, where $\overline{\text{verb}}$ is a tensor of order 3 and $\times$ is tensor contraction:

\begin{equation*}
\ov{svo} = (\overline{\textmd{verb}} \times \overrightarrow{\textmd{obj}}) \times \overrightarrow{\textmd{subj}}
\end{equation*}

Finally, phrase and sentence vectors have been also produced by the application of neural architectures, such as recursive or recurrent neural networks \cite{socher2012,cheng2015} and convolutional neural networks \cite{nal2014}. These models have been shown to perform well on large scale entailment tasks such as the ones introduced by the RTE challenge. 

\section{Distributional inclusion hypothesis}
\label{sec:DIH}

The distributional inclusion hypothesis (DIH) \cite{Dagan1999,geffet2005,herbelot2013} is based on the fact that whenever a word $u$ entails a word $v$, then it makes sense to replace  instances of   $u$ with   $v$. For example,  `cat' entails `animal', hence in the sentence `a cat is asleep', it makes sense to replace `cat' with `animal' and obtain `an animal is asleep'.  On the other hand, `cat' does not entail `butterfly', and indeed it does not make sense to do a similar substitution and obtain  the sentence  `a butterfly is asleep'.  This hypothesis has  limitations, the main one being that it only makes sense in upward monotone contexts. For instance, the substitution of $u$ for $v$ would not work for sentences that have negations or  quantifiers such as `all' and `none'. As a result, one cannot replace `cat' with `animal' in  sentences such as  `all cats are asleep' or `a cat is not asleep'. Despite this and other limitations, the DIH has been subject to a good amount of study in the distributional semantics community and its predictions have been validated \cite{geffet2005,kotlerman2010}. 

Formally, the DIH says that if word $u$ entails word $v$,  then the set of features of $u$ are included in the set of features of $v$. In the context of a distributional model of meaning, the term \textit{feature} refers to a non-zero dimension of the distributional vector of a word.  By denoting  the features of a distributional vector $\ov{v}$ by  $\F(\ov{v})$, one can  symbolically express the DIH as follows:

\begin{equation}
u \vdash v  \quad \text{whenever}   \quad \F(\ov{u}) \subseteq \F(\ov{v})
\label{eq:featureincl}
\end{equation}

The research on the DIH can be categorised into two classes. In the first class, the degree of entailment between two words is based on the distance between the vector representations of the words. This distance must be measured by asymmetric means, since entailment is directional. Examples of measures used here are entropy-based measures such as KL-divergence \cite{Chen1996}. KL-divergence is only defined when the support of  $\ov{v}$ is included in the support of  $\ov{u}$. To overcome this restriction, a variant referred to by $\alpha$-skew \cite{Lee1999} has been proposed (for $\alpha \in (0,1]$  a smoothing parameter). {\em Representativeness} provides another way of smoothing the KL-divergence. The formulae for these are as follows, where abusing the notation we take $\ov{u}$ and $\ov{v}$ to also denote the probability distributions of $u$ and $v$:

\vspace{-0.4cm}
\begin{gather*}
   D_{\rm KL} (\ov{v} \| \ov{u}) = \sum_i v_i (\ln v_i - \ln u_i) 
\quad\quad
   s_\alpha(\ov{u},\ov{v}) = D_{\rm KL}(\ov{v}\| \alpha \ov{u}+(1-\alpha)\ov{v}) \\
    R_{{\rm D}}(\ov{v}\|\ov{u}) = \frac{1}{1+D_{\rm KL}(\ov{v}||\ov{u})}
\end{gather*}

Representativeness turns KL-divergence into a number in the unit interval $[0,1]$. As a result we obtain  $0\leq R_{{\rm D}}(\ov{v}\|\ov{u})  \leq 1$, with $R_{{\rm D}}(\ov{v}\|\ov{u})  = 0$ when the support of $\ov{v}$ is not included in the support of $\ov{u}$  and $R_{{\rm D}}(\ov{v}\|\ov{u})  = 1$, when $\ov{u}$ and $\ov{v}$ represent the same distribution. 



The research done in the second class attempts a more direct measurement of the inclusion of features, with the simplest possible case returning a binary value for inclusion or lack thereof. Measures developed by \newcite{weeds2004} and \newcite{clarke2009} advance this simple methods by arguing that not all features  play an equal role in representing words and hence they should not be treated equally  when it comes to measuring entailment. Some features are more ``pertinent'' than others and these features have to be given a higher weight when computing inclusion. For example, `cat' can have a non-zero coordinate on all of the features `mammal, miaow, eat, drink, sleep'. But the amount of these coordinates differ, and one can say that, for example,  the higher the coordinate the more pertinent the feature.  Pertinence is computed by various different measures, the most recent of which is {\em balAPinc} \cite{kotlerman2010}, where {\em LIN} is Lin's similarity \cite{lin1998} and {\em APinc} is an asymmetric measure:

\[
 \textit{balAPinc}(u,v) = \sqrt{\textit{LIN}(u,v) \cdot \textit{APinc}(u,v)}
\qquad   \textit{APinc}(u,v) = \frac{\sum_r \left[ P(r) \cdot \textit{rel}'(f_r)\right]}{|\F(\ov{u})|}
\]
{\em APinc} applies the DIH via the idea that features with high values in $\F(\ov{u})$ must also have high values in $\F(\ov{v})$. In the above formula, $f_r$ is the feature in $\F(\ov{u})$ with rank $r$; $P(r)$ is the precision at rank $r$; and $\textit{rel}'(f_r)$ is a weight computed as follows:

\begin{equation}
  rel'(f) = \left\{
  \begin{array}{lr}
     1-\frac{\textit{rank}(f,\F(\ov{v}))}{|\F(\ov{v})|+1} & f \in \F(\ov{v}) \\
     0 & o.w.
  \end{array} 
  \right.
\end{equation}

\noindent
where $\textit{rank}(f,\F(\ov{v}))$ shows the rank of feature $f$ within the entailed vector. In general, \textit{APinc} can be seen as a version of average precision that reflects lexical inclusion.

\section{Measuring feature inclusion at the phrase/sentence level}

In recent work, the authors of this paper introduced a variation of the {\em APinc} and {\em balAPinc} measures aiming to address the extra complications imposed when evaluating entailment at the phrase/sentence level \cite{lacl2016}. The modified measures differ from the original ones in two aspects. Firstly, in a compositional distributional model, the practice of considering only non-zero elements of the vectors as features becomes too restrictive and thus suboptimal for evaluating entailment; indeed, depending on the form of the vector space and the applied compositional operator (especially in intersective models, see Sections \ref{sec:compdistDIH} and \ref{sec:tensors}), an element can get very low values without however ever reaching zero. The new measures exploit this blurring of the notion of ``featureness'' to the limit, by letting $\F(\ov{w})$ to include   all the dimensions of $\ov{w}$.

Secondly, the continuous nature of distributional models is further exploited by providing a stronger realization of the idea that $u\vdash v$ whenever $v$ occurs in all the contexts of $u$. Let $f_r^{(u)}$ be a feature in $\F(\ov{u})$ with rank $r$ and $f_r^{(v)}$ the corresponding feature in $\F(\ov{v})$, we remind that Kotlerman et al. consider that feature inclusion holds at rank $r$ whenever $f_r^{(u)} > 0$ and $f_r^{(v)} > 0$; the new measures strengthen this assumption by requiring that $f_r^{(u)} \leq f_r^{(v)}$. Incorporating these modifications in the {\em APinc} measure, $P(r)$ and $rel'(f_r)$ are redefined as follows:

\vspace{-0.2cm}
\begin{gather}
  P(r) = \frac{\big|\{ f_r^{(u)} | f_r^{(u)} \leq f_r^{(v)}, 0 < r \leq |\ov{u}| \}\big|}{r} ~~~~~~~~~~~~~~~~~
  rel'(f_r) = \left\{ 
     \begin{array}{lr}
        1 & f_r^{(u)} \leq f_r^{(v)} \\
        0 & o.w.
     \end{array}
     \right.
     \label{equ:rel-sent}
\end{gather}  

The new relevance function subsumes the old one, as  by definition high-valued features in $\F(\ov{u})$ must be even higher in $\F(\ov{v})$. The new {\em APinc} at the phrase/sentence level thus becomes as follows:

\begin{equation}
 \textit{SAPinc}(u,v) = \frac{\sum_r \left[ P(r) \cdot \textit{rel}'(f_r)\right]}{|\ov{u}|}
\end{equation} 

\noindent where $|\ov{u}|$ is the number of dimensions of $\ov{u}$. Further, we notice that {\em balAPinc} is the geometric average of {\em APinc}  with  Lin's similarity measure, which is symmetric. According to   \cite{kotlerman2010}, the  rationale of including a symmetric measure was that {\em APinc} tends to return unjustifyingly high scores when the entailing word is infrequent, that is, when the feature vector of the entailing word is very short; the purpose of the symmetric measure was to penalize the result, since in this case the similarity of the narrower term with the broader one is usually low. However, now that all feature vectors have the same length, such a balancing action is unnecessary;  more importantly, it introduces a strong element of symmetry in a measure that is intended to be strongly asymmetric. We cope with this issue by replacing Lin's measure with representativeness on KL-divergence,\footnote{Using other asymmetric measures is also possible; the choice of representativeness on KL-divergence was based on informal experimentation which showed that this combination works better than other options in practice.}  obtaining the following new version of {\em balAPinc}:

\begin{equation}
  \textit{SBalAPinc}(u,v) = 
  \sqrt{R_{\rm D}(\ov{u}\|\ov{v}) \cdot \textit{SAPinc}(\ov{u},\ov{v})}
\end{equation}

Recall that $R_{\rm D}(p\|q)$ is asymmetric, measuring the extent to which $q$ represents (i.e. is similar to) $p$. So the term $R_{\rm D}(\ov{u}\|\ov{v})$ in the above formula measures how well the {\em broader} term $v$ represents the narrower one $u$; as an example, we can think that the term `animal' is representative of `cat', while the reverse is not true. The new measure aims at: (i) retaining a strongly asymmetric nature; and (ii) providing a more fine-grained element of entailment evaluation. 

\section{Generic feature inclusion in  compositional models}
\label{sec:compdistDIH}

In the presence of a compositional operator, features of phrases or sentences adhere to set-theoretic properties. For simple additive and multiplicative models, the set of features of a phrase/sentence is  derived from the set of features of their words using  union and intersection. It is slightly less apparent (and for reasons of space we will not give details here) that the features of point-wise minimum and maximum of  vectors  are also derived from the intersection and union of their features, respectively. That is:

\vspace{-0.4cm}
\begin{gather*}
{\cal F}(\ov{v_1} + \cdots + \ov{v_n}) = {\cal F}(\max(\ov{v_1}, \cdots,  \ov{v_n})) = {\cal F}(\ov{v_1}) \cup \cdots \cup  {\cal F}(\ov{v_n}) \\
{\cal F}(\ov{v_1} \odot \cdots \odot \ov{v_n}) = {\cal F}(\min(\ov{v_1}, \cdots,  \ov{v_n})) = {\cal F}(\ov{v_1}) \cap \cdots \cap  {\cal F}(\ov{v_n})
\end{gather*}
\vspace{-0.4cm}

As shown in \cite{lacl2016}, element-wise composition of this form lifts naturally from the word level to phrase/sentence level; specifically, for two sentences $s_1=u_1\dots u_n$ and $s_2=v_1\dots v_n$ for which $u_i\vdash v_i~\forall i \in [1,n]$, it is always the case that $s_1\vdash s_2$. This kind of lifting of the entailment relationship from words to the phrase/sentence level also holds for tensor-based models \cite{isaim2016}. 

In general, feature inclusion is a more complicated process for tensor-based settings, since in this case the composition operation is matrix multiplication and tensor contraction. As an example, consider the simple case of a matrix multiplication between a $m \times n$ matrix $\mathbf{M}$ and a $n \times 1$ vector $\ov{v}$. Matrix $\mathbf{M}$ can be seen as a list of column vectors $(\ov{w_1},   \ov{w_2},  \cdots,  \ov{w_n})$, where  $\ov{w_i}=(w_{1i},\cdots,w_{mi})^\mathsf{T}$.  The  result of the matrix multiplication is a  combination of scalar multiplications of each element $v_i$  of the vector $\ov{v}$ with the corresponding column vector $\ov{w_i}$ of the matrix $\mathbf{M}$. That is, we have:

\[
\left(\begin{array}{ccc}
w_{11} & \cdots & w_{1n}\\
w_{21} & \cdots & w_{2n}\\
\vdots & & \vdots\\
w_{m1} & \cdots & w_{mn}
\end{array}\right)
\times
\left(
\begin{array}{c}
v_1\\
 \vdots \\ 
 v_n
\end{array}\right) 
 =
  v_1  \ov{w_1} + 
v_2  \ov{w_2} + 
\cdots  + 
v_n   \ov{w_n}
\]

Looking at  $\mathbf{M} \times \ov{v}$ in this way  enables us to describe ${\cal F}(\mathbf{M} \times \ov{v})$ in terms of  the union of ${\cal F}(\ov{w_i})$'s where $v_i$ is non zero, that is, we have:

\vspace{-0.3cm}
\[{\cal F}(\mathbf{M} \times \ov{v} ) = \bigcup_{v_i \neq 0} {\cal F}(\ov{w_i})
\]
By denoting  $v_i$ a feature whenever it is non-zero, we obtain an equivalent form as follows: 

\vspace{-0.3cm}
\begin{equation}
\bigcup_i  {\cal F}(\ov{w_i})  \mid_{{\cal F}(v_i)}
\label{equ:fincl}
\end{equation}

The above notation says that we collect features of each $\ov{w_i}$ vector but only up to ``featureness'' of $v_i$, that is up to $v_i$ being non-zero.  This can be extended to tensors of higher order; a tensor of order 3, for example, can be seen as a list of matrices, a tensor of order 4 as a list of ``cubes'' and so on. For the case of this paper, we will not go beyond matrix multiplication and cube contraction. 

\section{Feature inclusion in concrete constructions of tensor-based models}
\label{sec:tensors}

While the previous section provided a generic analysis of the feature inclusion behaviour of tensor-based models, the exact feature inclusion properties of these models depend on the specific concrete constructions, and in principle get a form more refined than that of simple intersective or union-based composition. In this section we investigate a number of tensor-based models with regard to feature inclusion, and derive their properties.

\subsection{Relational model}
\label{sec:rel}

As a starting point we will use the model of \newcite{grefenstette2011}, which adopts an extensional approach and builds the tensor of a relational word from the vectors of its arguments. More specifically, the tensors for adjectives, intransitive verbs, and transitive verbs are defined as below, respectively:

\vspace{-0.2cm}
\begin{equation}
  \ovl{adj} = \sum_i \ov{Noun}_i~~~~~~~~
  \ovl{verb}_{\rm IN} = \sum_i \ov{Sbj}_i~~~~~~~~
  \ovl{verb}_{\rm TR} =  \sum_i \ov{Sbj}_i \otimes \ov{Obj}_i
  \label{equ:gs2011}
\end{equation}

\noindent where $\ov{Noun_i}$, $\ov{Sbj}_i$ and $\ov{Obj_i}$ refer to the distributional vectors of the nouns, subjects and objects that occurred as arguments for the adjective and the verb across the training corpus. For the case of a subject-verb sentence and a verb-object phrase, composition reduces to element-wise multiplication of the two vectors, and the features of the resulting sentence/phrase vector get the following form (with $\ov{s}$ and $\ov{o}$ to denote the vectors of the subject/verb of the phrase/sentence):

\begin{equation}
   \F(\ov{sv}) = \bigcup_i \F(\ov{Sbj}_i) \cap \F(\ov{s})~~~~~~~~~~~~
   \F(\ov{vo}) = \bigcup_i  \F(\ov{Obj}_i)\cap \F(\ov{o})
\end{equation}

For a transitive sentence, the model of \newcite{grefenstette2011} returns a matrix, computed in the following way:

\vspace{-0.2cm}
\[
   \ovl{svo}_{\rm Rel} = \ovl{verb} \odot (\ov{s}\otimes \ov{o})
\]

\noindent where $\ovl{verb}$ is defined as in Equation \ref{equ:gs2011}. By noticing that $\F(\ov{u} \otimes \ov{v}) = \F(\ov{u}) \times \F(\ov{v})$, with symbol $\times$ to denote in this case the {\em cartesian product} of the two feature sets, we define the feature set of a transitive sentence as follows:

\vspace{-0.2cm}
\begin{equation}
  \F(\ovl{svo}_{\rm Rel}) = \bigcup_i \F(\ov{Sbj}_i) \times \F(\ov{Obj}_i) \cap \F(\ov{s}) \times \F(\ov{o})
  \label{equ:rel}
\end{equation}

Equation \ref{equ:rel} shows that the features of this model are pairs $(f_{sbj},f_{obj})$, with $f_{sbj}$ a subject-related feature and $f_{obj}$ an object-related feature, providing a fine-grained representation of the sentence. Throughout this paper, we refer to this model as {\em relational}.

\subsection{Frobenius models}
\label{sec:frob}

As pointed out in \cite{KartSadrPul-COLING-2013}, the disadvantage of the relational model is that their resulting representations of verbs have one dimension less than what their types dictate. According to the type assignments, an intransitive verb has to be a matrix and a transitive verb a cube, where as in the above we have a vector and a matrix. A solution presented in \cite{KartSadrPul-COLING-2013} suggested the use of \emph{Frobenius} operators in order to expand vectors and matrices into higher order tensors by embedding them into the the diagonals. For example, a vector is embedded into a matrix by putting it in the diagonal of a matrix and padding the off-diagonal elements with zeros. Similarly, one can embed a matrix into a cube by putting it into the main diagonal and pad the rest with zeros. Using this method, for example, one could transform a simple intersective model in tensor form by embedding the context vector of a verb $\ov{v}$ first into a matrix and then into a cube. For a transitive sentence, one could use the matrix defined in Equation \ref{equ:gs2011} and derive a vector for the meaning of the sentence in two ways, each one corresponding to a different embedding of the matrix into a tensor of order 3:

\begin{equation}
   \ov{svo}_{\rm CpSbj} = \ov{s} \odot (\ovl{verb} \times \ov{o})~~~~~~~~~~~~~
   \ov{svo}_{\rm CpObj} = (\ov{s}^\mathsf{T} \times \ovl{verb}) \odot \ov{o}
\end{equation}

We refer to these models as Copy-Subject and Copy-Object, correspondingly. In order to derive their feature inclusion properties, we first examine the form of the sentence vector produced when the verb is composed with a new subject/object pair:

\begin{gather*}
   \ov{svo}_{\rm CpSbj} = \ov{s} \odot (\ovl{verb} \times \ov{o}) = \ov{s} \odot \sum_i \ov{Sbj_i} \langle \ov{Obj}_i | \ov{o} \rangle \\
   \ov{svo}_{\rm CpObj} = (\ov{s}^\mathsf{T} \times \ovl{verb}) \odot \ov{o} = \ov{o} \odot \sum_i \ov{Obj_i} \langle \ov{s} | \ov{Sbj}_i \rangle 
\end{gather*}

We can now define the feature sets of the two models using notation similar to that of Equation \ref{equ:fincl}:

\begin{gather}
\begin{split}
  \F(\ov{svo}_{\rm CpSbj}) = \F(\ov{s}) \cap \bigcup_i  \F(\ov{Sbj}_i) \mid_{\F\left(\langle \ov{Obj}_i | \ov{o} \rangle\right)} \\
  \F(\ov{svo}_{\rm CpObj}) = \F(\ov{o}) \cap \bigcup_i  \F(\ov{Obj}_i) \mid_{\F\left(\langle \ov{s} | \ov{Sbj}_i \rangle\right)}
\end{split}
\label{equ:frob}
\end{gather}

The symbol $|$ defines a restriction on feature inclusion based on how well the arguments of the sentence fit to the arguments of the verb. For a subject-object pair $(Sbj,Obj)$ that has occured with the verb in the corpus, this translates to the following:

\begin{itemize}
  \item {\em Copy-Subject:} Include the features of $Sbj$ up to similarity of $Obj$ with the sentence object
  \item {\em Copy-Object:} Include the features of $Obj$ up to similarity of $Sbj$ with the sentence subject
\end{itemize}  

Note that each  of the Frobenius models puts emphasis on a different argument of a sentence; the Copy-Subject model collects features of the subjects that occured with the verb, while the Copy-Object model collects features from the verb objects. It is reasonable then to further combine the two models in order to get a more complete representation of the sentence meaning, and hence its feature inclusion properties. Below we define the feature sets of two variations, where this combination is achieved via addition (we refer to this model as {\em Frobenius additive}) and element-wise multiplication ({\em Frobenius multiplicative}) of the vectors produced by the individual models  \cite{kartsadrqpl2014}:

\vspace{-0.2cm}
\begin{gather}
\begin{split}
  \F(\ov{svo}_{\rm FrAdd}) = \F(\ov{svo}_{\rm CpSbj}) \cup \F(\ov{svo}_{\rm CpObj}) \\
  \F(\ov{svo}_{\rm FrMul}) = \F(\ov{svo}_{\rm CpSbj}) \cap \F(\ov{svo}_{\rm CpObj})
\end{split}  
\end{gather}

\noindent where $\F(\ov{svo}_{\rm CpSbj})$ and  $\F(\ov{svo}_{\rm CpObj})$ are defined as in Equation \ref{equ:frob}.

\subsection{Projective models}
\label{sec:proj}

In this section we provide an alternative solution and remedy the problem of having lower dimensions than the required by arguing that the sentence/phrase space should be spanned by the vectors of the arguments of the verb across the corpus. Thus we create verb matrices for intransitive sentences and verb phrases by summing up {\em projectors} of the argument vectors, in the following way:

\begin{equation}
  \ovl{v}_{\it itv}:=  \sum_i \ov{Sbj_i} \otimes \ov{Sbj_i}~~~~~~~~~~~~~
  \ovl{v}_{\it vp} := \sum_i \ov{Obj_i} \otimes \ov{Obj_i}
  \label{equ:proj}
\end{equation}

When these verbs are composed with some subject/object to form a phrase/sentence, each vector in the spanning space is weighted by its similarity (assuming normalized vectors) with the vector of that subject/object, that is:

\begin{gather}
  \ov{sv}_{\rm Prj} = \ov{s}^{\mathsf{T}} \times \overline{v}_{itv} = \sum_i \langle \ov{Sbj_i}| \ov{s}\rangle \ov{Sbj_i}~~~~~~~~~~~
\ov{vo}_{\rm Prj} = \overline{v}_{vp} \times \ov{o} = \sum_i \langle \ov{Obj_i}|\ov{o}\rangle \ov{Obj_i}  
\label{equ:proj-sent}
\end{gather}

Translating the above equations to feature inclusion representations will give:

\begin{equation}
    \F(\ov{sv}_{\rm Prj}) = \bigcup_i {\cal F}(\ov{Sbj}_i) \mid_{\F\left(\langle \ov{Sbj_i}|\ov{s}\rangle\right)}~~~~~~~~~~~~~~
    \F(\ov{vo}_{\rm Prj}) =  \bigcup_i {\cal F}(\ov{Obj}_i) \mid_{\F\left(\langle \ov{Obj_i}|\ov{o}\rangle\right)}
\end{equation}

\noindent with symbol $|$ to define again a restriction on feature inclusion based on the similarity of the arguments with the subject or object of the sentence/phrase. For the subject-verb case, this reads: ``include a subject that occured with the verb, up to its similarity with the subject of the sentence''. For the  case of a transitive verb (a function of two arguments), we define the sentence space to be spanned by the average of the argument vectors,  obtaining:

\begin{equation*}
    \ovl{v}_{trv} := \sum_{i} \ov{Sbj_i} \otimes \left( \frac{\ov{Sbj_i} + \ov{Obj_i}}{2} \right) \otimes \ov{Obj_i}
\end{equation*}

The meaning of a transitive sentence then is computed as:

\begin{equation}
\label{equ:tran-cat}
\ov{s v o}_{\rm Prj} = \ov{s}^{\mathsf{T}} \times \ovl{v}_{trv} \times \ov{o} =  \sum_i  \left[ \langle \ov{s} \mid \ov{Sbj_i}\rangle \langle \ov{Obj_i} \mid \ov{o} \rangle    \left(\frac{\ov{Sbj_i} + \ov{Obj_i}}{2}\right) \right]
\end{equation}

Feature-wise, the above translates to the following:

\begin{equation}
   \F(\ov{svo}_{\rm Prj}) = \bigcup_i \left(\F(\ov{Sbj}_i) \cup \F(\ov{Obj_i}) \right) 
\mid_{\F\left(\langle \ov{s}|\ov{\textit{Sbj}}_i\rangle\right) \F\left(\langle \ov{\textit{Obj}}_i|\ov{o}\rangle\right)}
\end{equation}

Note that in contrast with the relational and Frobenius models, which all employ an element of intersective feature selection, the projective models presented in this section are purely union-based. 

\subsection{Inclusion of verb vectors}
\label{sec:verbincl}

The models of the previous sections provide a variety of options for representing the meaning of a verb from its arguments. However, none of these constructions takes into account the distributional vector of the verb itself, which includes valuable information that could further help in entailment tasks. We remedy this problem by embedding the missing information into the existing tensors; for example, we can amend the tensors of the projective model as follows:

\vspace{-0.2cm}
\begin{equation}
  \ovl{v}_{itv} = \sum_i \ov{Sbj_i} \otimes \left(\ov{Sbj_i}\odot \ov{v}\right)~~~~~~~~~~~~~
  \ovl{v}_{vp} = \sum_i \left(\ov{Obj_i}\odot \ov{v}\right) \otimes \ov{Obj_i}
\end{equation}

\noindent with $\ov{v}$ denoting the distributional vector of the verb. In the context of an intransitive sentence, now we have the following interaction:

\vspace{-0.2cm}
\begin{equation}
   \ov{sv} =  \ov{s}^{\mathsf{T}} \times \ovl{v}_{itv} = \ov{s}^{\mathsf{T}} \times  \sum_i \ov{Sbj_i} \otimes \left(\ov{Sbj_i}\odot \ov{v}\right) =
    \ov{v} \odot \sum_i \langle \ov{s}|\ov{Sbj_i}\rangle \ov{Sbj_i}
\end{equation}

We see that the result of the standard projective model (Equation \ref{equ:proj-sent}) is now enchanced with an additional step of interesective feature selection. In feature inclusion terms, we get:

\vspace{-0.2cm}
\begin{equation}
    \F(\ov{sv}) = \F(\ov{v}) \cap \F(\ov{sv}_{\rm Prj}) ~~~~~~~~~~~~~~
    \F(\ov{vo}) =  \F(\ov{v}) \cap \F(\ov{vo}_{\rm Prj})
\end{equation}

It is easy to show that similar formulae hold for the relational and Frobenius models. 




\section{Experimental setting}
\label{sec:exp}

We evaluate the feature inclusion behaviour of the tensor-based models of Section \ref{sec:tensors}  in three different tasks; specifically, we measure upward-monotone entailment between (a) intransitive sentences; (b) verb phrases; and (c) transitive sentences. We use the entailment datasets introduced in \cite{lacl2016}, which consist of 135 subject-verb pairs, 218 verb-object pairs, and 70 subject-verb-object pairs, the phrases/sentences of which stand in a fairly clear entailment relationship. Each dataset has been created using hypernym-hyponym relationships from WordNet, and it was extended with the reverse direction of the entailments as negative examples, creating three strictly directional entailment datasets of 270 (subject-verb), 436 (verb-object) and 140 (subject-verb-object) entries. Some examples of positive entailments from all three categories include:\footnote{The datasets are available at {\tt http://compling.eecs.qmul.ac.uk/resources/}.}

\scriptsize
\begin{center}
\begin{tabular}{c|c|c}

{\bf Subject-verb} & {\bf Verb-object} & {\bf Subject-verb-object} \\
\hline
evidence suggests $\vdash$ information expresses & sign contract $\vdash$ write agreement & book presents account $\vdash$ work shows evidence  \\
survey reveals $\vdash$ work shows                           & publish book $\vdash$ produce publication & woman marries man $\vdash$ female joins male  \\
player plays $\vdash$ contestant compete                & sing song $\vdash$ perform music & author retains house $\vdash$ person holds property \\
study demonstrates $\vdash$ examination shows   & reduce number $\vdash$ decrease amount & study demonstrates importance $\vdash$ work shows value \\
summer finishes $\vdash$ season ends                    & promote development $\vdash$ support event & experiment tests hypothesis $\vdash$ research evaluates proposal  \vspace{0.3cm} 
\end{tabular}
\end{center}
\normalsize

In all cases, we first apply a compositional model to the phrases/sentences of each pair in order to create vectors representing their meaning, and then we evaluate the entailment relation between the phrases/sentences by using these composite vectors as input to a number of entailment measures. The goal is to see which combination of compositional model/entailment measure is capable of better recognizing strictly directional entailment relationships between phrases and sentences. 

We experimented with a variety of entailment measures, including {\em SAPinc} and {\em SBalAPinc} as in \cite{lacl2016}, their word-level counterparts \cite{kotlerman2010}, KL-divergence (applied to smoothed vectors as in \newcite{Chen1996}), $\alpha$-skew  with  $\alpha=0.99$ as in \newcite{kotlerman2010},  {\em WeedsPrec} as in \newcite{weeds2004}, and  {\em ClarkeDE} as in \newcite{clarke2009}. We  use strict feature inclusion as a baseline; in this case, entailment holds only when $\F(\ov{\textit{phrase}}_1) \subseteq \F(\ov{\textit{phrase}}_2)$. For compositional operators, we experimented with element-wise vector multiplication and MIN, vector addition and MAX, and the tensor models presented in Section \ref{sec:tensors}. Informal experimentation showed that directly embedding distributional information from verb vectors in the tensors (Section \ref{sec:verbincl}) works considerably better than the simple versions, so the results we report here are based on this approach. We also present results for a least squares fitting model, which approximates the distributional behaviour of holistic phrase/sentence vectors along the lines of  \cite{baroni2010}. Specifically, for each verb,  we compute  an estimator that predicts the $i$th element of the resulting vector as follows:

\vspace{-0.2cm}
\begin{equation*}
  \ov{w_i} = (\mathbf{X}^{\mathsf{T}} \mathbf{X})^{-1} \mathbf{X}^{\mathsf{T}} \ov{y_i}
\end{equation*}

Here, the rows of matrix $\mathbf{X}$ are the vectors of the subjects (or objects) that occur with our verb, and $\ov{y_i}$ is a vector containing the $i$th elements of the holistic phrase vectors across all training instances; the resulting $\ov{w_i}$'s form the rows of our verb matrix. This model could be only implemented for  verb-object and subject-verb phrases due to data sparsity problems. As our focus is on analytic properties of features, we did not experiment with any neural  model. 
 
Regarding evaluation, since the tasks follow a binary classification objective and our models return a continuous value, we report {\em area under curve} (AUC). This reflects the generic discriminating power of a binary classifier by evaluating the task at every possible threshold. In all the experiments,  we used a 300-dimensional PPMI vector space trained on the concatenation of UKWAC and Wikipedia corpora. The context was defined as a 5-word window around the target word.

\section{Results and discussion}
\label{sec:results}

\begin{table}[h!]
\begin{center}
\scriptsize
\begin{tabular}{|c||c|cccccc|cc|}
\hline
 \multicolumn{10}{|c|}{\bf Subject-verb task} \\
\hline
{\bf Model} & Inclusion  & KL-div & $\alpha$Skew & WeedsPrec & ClarkeDE & APinc & balAPinc & SAPinc & SBalAPinc \\
\hline\hline
 Verb                      & 0.59 & 0.59 & 0.63 & 0.67 & 0.57 & 0.69 & 0.65 & 0.65 & 0.65 \\
 \hline
 $\odot$                   & 0.54 & 0.66 & 0.75 & 0.75 & 0.66 & 0.78 & 0.72 & {\bf 0.81} & {\bf 0.81} \\
 ${\rm min}$               & 0.54 & 0.68 & 0.72 & 0.75 & 0.63 & 0.78 & 0.71 & 0.74 & 0.75 \\
 $+$                       & 0.63 & 0.57 & 0.74 & 0.65 & 0.62 & 0.72 & 0.70 & 0.72 & 0.72 \\

 ${\rm max}$ & 0.63 & 0.57 & 0.70 & 0.65 & 0.60 & 0.71 & 0.65 & 0.71 & 0.71 \\
 \hline
 Least-Sqr                 & 0.50 & 0.59 & 0.62 & 0.59 & 0.56 & 0.60 & 0.58 & 0.63 & 0.64 \\
 $\otimes_{\rm proj}$ & 0.59 & 0.59 & 0.65 & 0.67 & 0.59 & 0.70 & 0.67 & 0.71 & 0.69 \\
 $\otimes_{\rm rel/frob}$ & 0.54 & 0.64 & 0.77 & 0.74 & 0.68 & 0.78 & 0.73 &     {\bf 0.84} & {\bf 0.83} \\ 
 \hline\hline
 \multicolumn{10}{|c|}{\bf Verb-object task} \\
 \hline
{\bf Model} & Inclusion  & KL-div & $\alpha$Skew & WeedsPrec & ClarkeDE & APinc & balAPinc & SAPinc & SBalAPinc \\
\hline\hline

 Verb                      & 0.58 & 0.62 & 0.65 & 0.67 & 0.58 & 0.69 & 0.66 & 0.62 & 0.66 \\
 \hline
 $\odot$                   & 0.52 & 0.64 & 0.74 & 0.70 & 0.67 & 0.75 & 0.70 & {\bf 0.82} & {\bf 0.79} \\
 ${\rm min}$               & 0.52 & 0.66 & 0.70 & 0.71 & 0.63 & 0.75 & 0.69 & 0.74 & 0.74 \\
 $+$                       & 0.64 & 0.61 & 0.75 & 0.68 & 0.63 & 0.74 & 0.71 & 0.72 & 0.73  \\
 ${\rm max}$  & 0.64 & 0.62 & 0.73 & 0.68 & 0.62 & 0.72 & 0.68 & 0.62 & 0.66 \\
  \hline
 Least-Sqr                 & 0.50 & 0.58 & 0.57 & 0.56 & 0.53 & 0.56 & 0.55 & 0.58 & 0.59 \\
 $\otimes_{\rm proj}$ & 0.58 & 0.60 & 0.66 & 0.67 & 0.60 & 0.70 & 0.67 & 0.68 & 0.68 \\
 $\otimes_{\rm rel/frob}$ & 0.52 & 0.63 & 0.75 & 0.71 & 0.67 & 0.75 & 0.70 & {\bf 0.82} & {\bf 0.79} \\
 \hline
 
\end{tabular}
\end{center}
\caption{AUC results for the subject-verb and verb-object tasks. `Verb' refers to a non-compositional baseline, where the vector/tensor of the phrase is taken to be the vector/tensor of the head verb. $\odot$, $+$ refer to vector element-wise multiplication and addition, respectively, $\otimes_{\rm proj}$ to the projective tensor models of Section \ref{sec:proj}, and $\otimes_{\rm rel/frob}$ to the construction of Section \ref{sec:rel}. The tensor models (except the least squares one) are further enhanced with information from the distributional vector of the verb, as detailed in Section \ref{sec:verbincl}.}
\label{tbl:intr}
\end{table}

\begin{table}[h!]
\begin{center}
\scriptsize
\begin{tabular}{|c||c|cccccc|cc|c|}
\hline
{\bf Model} & Inclusion  & KL-div & $\alpha$Skew & WeedsPrec & ClarkeDE & APinc & balAPinc & SAPinc & SBalAPinc \\
\hline\hline

 Verb                      & 0.61 & 0.61 & 0.66 & 0.69 & 0.58 & 0.74 & 0.67 & 0.59 & 0.63  \\
 \hline
 $\odot$                   & 0.55 & 0.65 & 0.74 & 0.79 & 0.67 & 0.76 & 0.71 & {\bf 0.80} & {\bf 0.80} \\
 ${\rm min}$               & 0.55 & 0.71 & 0.74 & 0.78 & 0.63 & 0.77 & 0.71 & 0.73 & 0.76  \\ 
 $+$                       & 0.58 & 0.54 & 0.71 & 0.59 & 0.60 & 0.65 & 0.64 & 0.67 & 0.67  \\
 ${\rm max}$               & 0.58 & 0.55 & 0.68 & 0.58 & 0.58 & 0.63 & 0.61 & 0.60 & 0.61  \\  
 \hline
 Least-Sqr                 & -- & -- & -- & -- & -- & -- & -- & -- & -- \\
 $\otimes_{\rm rel}$       & 0.51 & 0.64 & 0.78 & 0.79 & 0.69 & 0.79 & 0.72 & 0.84 & 0.83 \\
 $\otimes_{\rm proj}$ & 0.64 & 0.60 & 0.70 & 0.69 & 0.61 & 0.74 & 0.70 & 0.75 & 0.76  \\
 $\otimes_{\rm CpSbj}$      & 0.57 & 0.65 & 0.73 & 0.77 & 0.63 & 0.73 & 0.68    & 0.79 & 0.78 \\
 $\otimes_{\rm CpObj}$      & 0.54 & 0.62 & 0.73 & 0.72 & 0.64 & 0.76 & 0.71 & 0.81 & 0.79 \\
 $\otimes_{\rm FrAdd}$      & 0.60 & 0.60 & 0.75 & 0.72 & 0.67 & 0.77 & 0.75 & 0.84 & 0.82 \\ 
 $\otimes_{\rm FrMul}$      & 0.55 & 0.62 & 0.76 & 0.81 & 0.68 & 0.78 & 0.73     & {\bf 0.86} & {\bf 0.83} \\ 
 \hline
\end{tabular}
\end{center}
\caption{AUC results for the subject-verb-object task. $\otimes_{\rm Rel}$ refers to the relational tensor model of Section \ref{sec:rel}, while $\otimes_{\rm CpSbj}$, $\otimes_{\rm CpObj}$, $\otimes_{\rm FrAdd}$, and $\otimes_{\rm FrMul}$ to the Frobenius models of Section \ref{sec:frob}. As in the other  tasks, the distributional vector of the verb has been taken into account in all tensor-based models except Least-Sqr.}
\label{tbl:svo}
\end{table}

The results are presented in Table \ref{tbl:intr} (subject-verb and verb-object task) and Table \ref{tbl:svo} (subject-verb-object task). In all cases, a combination of a Frobenius tensor model with one of the sentence-level measures ({\em SAPinc}) gives the highest performance. In general, {\em SAPinc} and {\em SBalAPinc} work very well with all the tested compositional models, achieving a cross-model performance higher than that of any other metric, for all three tasks. From a feature inclusion perspective, we see that models employing an element of interesective composition (vector multiplication, MIN, relational and Frobenius tensor models) have consistent high performances across all the tested measures. The reason may be that the intersective filtering avoids  generation of very dense vectors and thus facilitates entailment judgements based on the DIH. On the other hand, union-based compositional models, such as vector addition, MAX, and the projective tensor models, produce dense vectors for even very short sentences, which affects negatively the evaluation of entailment. The non-compositional verb-only baseline was worse than any compositional model other than the least-squares model, which is the only tensor model that did not perform well; this indicates that our algebraic tensor-based constructions  are  more robust against data sparsity problems than statistical models based on holistic vectors of phrases and sentences.


\section{Conclusion and future work}
\label{sec:conc}

In this paper we investigated  the application of the distributional inclusion hypothesis on evaluating entailment  between phrase and sentence vectors produced by compositional operators with a focus on tensor-based models. Our results showed that intersective composition in general, and the Frobenius tensor models in particular, achieve the best performance when evaluating upward monotone entailment, especially when combined with the sentence-level measures of \cite{lacl2016}. Experimenting with different versions of tensor models for entailment is an interesting topic that we plan to pursue further in a future paper. Furthermore, the extension of word-level entailment to phrases and sentences provides  connections with natural logic \cite{MacCartney2007}, a topic that is  worth a separate treatment and constitutes a future direction. 

\section*{Acknowledgments}

The authors gratefully acknowledge support by EPSRC for Career Acceleration Fellowship EP/\-J0\-02\-607/1 and AFOSR International Scientific Collaboration Grant FA9550-14-1-0079. 

\bibliography{coling2016-camera}
\bibliographystyle{acl}

\end{document}